\documentclass[sigconf]{acmart}
\AtBeginDocument{%
  \providecommand\BibTeX{{%
    \normalfont B\kern-0.5em{\scshape i\kern-0.25em b}\kern-0.8em\TeX}}}

\setcopyright{acmcopyright}
\copyrightyear{2018}
\acmYear{2018}
\acmDOI{XXXXXXX.XXXXXXX}

\acmConference[Conference acronym 'XX]{Make sure to enter the correct
  conference title from your rights confirmation emai}{June 03--05,
  2018}{Woodstock, NY}
\acmPrice{15.00}
\acmISBN{978-1-4503-XXXX-X/18/06}

\copyrightyear{2023}
\acmYear{2023}
\setcopyright{acmcopyright}\acmConference[WSDM '23]{Proceedings of the Sixteenth ACM International Conference on Web Search and Data Mining}{February 27-March 3, 2023}{Singapore, Singapore}
\acmBooktitle{Proceedings of the Sixteenth ACM International Conference on Web Search and Data Mining (WSDM '23), February 27-March 3, 2023, Singapore, Singapore}
\acmPrice{15.00}
\acmDOI{10.1145/3539597.3573035}
\acmISBN{978-1-4503-9407-9/23/02}

\begin{document}

%%
%% The "title" command has an optional parameter,
%% allowing the author to define a "short title" to be used in page headers.
\title{Unsupervised Question Duplicate and Related Questions Detection in e-learning platforms}

%%
%% The "author" command and its associated commands are used to define
%% the authors and their affiliations.
%% Of note is the shared affiliation of the first two authors, and the
%% "authornote" and "authornotemark" commands
%% used to denote shared contribution to the research.
\author{Maksimjeet Chowdhary}
\authornote{Both authors contributed equally to this research.}
\email{maksimjeet20566@iiitd.ac.in}
\affiliation{%
  \institution{IIIT Delhi}
  \city{Delhi}
  \country{India}}
% \orcid{1234-5678-9012}
\author{Sanyam Goyal}
\authornotemark[1]
\email{sanyam20116@iiitd.ac.in}
\affiliation{%
  \institution{IIIT Delhi}
%   \streetaddress{P.O. Box 1212}
%   \city{Dublin}
  \state{Delhi}
  \country{India}
  \postcode{43017-6221}
}

\author{Venktesh V}
\email{venkteshv@iiitd.ac.in}
\affiliation{%
  \institution{IIIT Delhi}
  \city{Delhi}
  \country{India}}

\author{Mukesh Mohania}
\email{mukesh@iiitd.ac.in}
\affiliation{%
  \institution{IIIT Delhi}
  \city{Delhi}
  \country{India}}

\author{Vikram Goyal}
\email{vikram@iiitd.ac.in}
\affiliation{%
  \institution{IIIT Delhi}
  \city{Delhi}
  \country{India}}

%%
%% By default, the full list of authors will be used in the page
%% headers. Often, this list is too long, and will overlap
%% other information printed in the page headers. This command allows
%% the author to define a more concise list
%% of authors' names for this purpose.
\renewcommand{\shortauthors}{Maksimjeet and Sanyam, et al.}

%%
%% The abstract is a short summary of the work to be presented in the
%% article.
\begin{abstract}
Online learning platforms provide diverse questions to gauge the learners' understanding of different concepts. The repository of questions has to be constantly updated to ensure a diverse pool of questions to conduct assessments for learners. However, it is impossible for the academician to manually skim through the large repository of questions to check for duplicates when onboarding new questions from external sources. Hence, we propose a tool \textit{QDup} in this paper that can surface near-duplicate and semantically related questions without any supervised data. The proposed tool follows an unsupervised hybrid pipeline of statistical and neural approaches for incorporating different nuances in similarity for the task of question duplicate detection. We demonstrate that \textit{QDup} can detect near-duplicate questions and also suggest related questions for practice with remarkable accuracy and speed from a large repository of questions. The demo video of the tool can be found at \url{https://www.youtube.com/watch?v=loh0_-7XLW4}.
\end{abstract}

%%
%% The code below is generated by the tool at http://dl.acm.org/ccs.cfm.
%% Please copy and paste the code instead of the example below.
%%


% \ccsdesc[500]{Computer systems organization~Embedded systems}
% \ccsdesc[300]{Computer systems organization~Redundancy}
% \ccsdesc{Computer systems organization~Robotics}
% \ccsdesc[100]{Networks~Network reliability}

%%
%% keyphrases. The author(s) should pick words that accurately describe
%% the work being presented. Separate the keyphrases with commas.

%%
%% This command processes the author and affiliation and title
%% information and builds the first part of the formatted document.
\maketitle

\section{Introduction}
The e-learning platforms usually curate a large repository of questions across subjects, chapters, and topics for conducting assessments to test the understanding of the learner. These repositories are constantly augmented with new questions. The new questions could be collected in batches from other platforms or external sources. They could also be added manually by the academicians. When new questions are added, there are cases of them being near-duplicates or related to existing questions in the data repository. It is impossible for the academicians to manually skim through the entire repository to check for duplicates. Hence, in this work, we propose a tool with support for bulk on-boarding of questions while surfacing duplicate questions already present in the database.

\begin{figure*}
    \centering
    \includegraphics[width=0.65\linewidth]{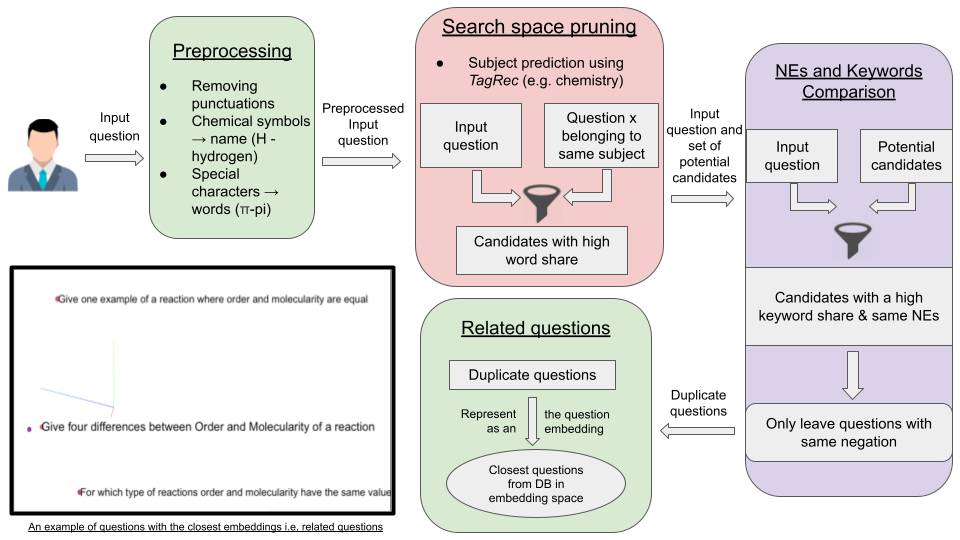}
    \caption{Duplicate Question Detection Pipleine}
    \label{fig:my_label}
\end{figure*}
The duplicate question detection task, particularly in the context of e-learning platforms, is a significant challenge due to the nature of the questions. Two questions can differ in entities or technical concepts though their verbiage and the rest of the semantics could be similar. In certain cases, though the questions are centered around the same entity and have mostly similar verbiage, the answers could be different. For example, the questions \textit{What is GDP?} and \textit{What is the significance of GDP?} might have high Jaccard or cosine similarity but are not duplicate questions. Hence, to encompass the mentioned scenarios, we define two questions to be duplicates of each other if they satisfy all of the following conditions:
\begin{itemize}
    \item The questions are lexically similar and have synonymous keyphrases or entities.
    \item The questions are semantically related.
    \item The correct answers to both the questions are equivalent
\end{itemize}

We also recommend related questions to aid the academicians in generating diverse questions for assessments. For instance, the questions \textit{What is the strongest bone in the body?} and \textit{What is the weakest bone in the body?} are related questions.

The duplicate text detection \cite{topic_model, cross_lang, Filice2018LearningPP,answer_info,das-etal-2016-together} is a well explored problem. These approaches range from comparing topics obtained through topic modelling \cite{topic_model}, comparing syntactic structure \cite{Filice2018LearningPP} to neural IR based methods \cite{das-etal-2016-together}. However, these approaches consider only uni-dimensional aspects of similarity, as mentioned earlier, and fail to identify duplicates in other scenarios where the questions only differ in entities. They also require significant amounts of the labeled dataset where questions are labeled as duplicates like CQADupStack \cite{cqadup,nakov-etal-2016-semeval-2016}, which is not available in the problem setting explained in this paper.

 The pipeline proposed in this paper is unsupervised and efficient in that it does not require any training. Since our approach is hybrid and uses a combination of classical and neural IR approaches, it is also efficient at inference time. The overview of the proposed pipeline can be seen in Figure 1. In summary, our core contributions are:
 
 \begin{itemize}
     \item We propose an unsupervised approach for near-duplicate detection in online learning platforms to enable smooth on-boarding of new questions. We also recommend related questions for serving diverse questions.
     \item We develop and release an easy-to-use tool that can support both individual and bulk on-boarding of questions at \\ \url{https://github.com/ADS-AI/QDup}.
 \end{itemize}

\section{System Design}
In this section, we describe the methodology used for searching for duplicate questions with respect to a large existing question repository. Given an input question $q_{new}=\{x_1,x_2...x_n\}$ of sequence length $n$ our goal is to surface exact duplicate questions $qdup_{exact}$, near-duplicates $qdup$ and related questions $q_{rel}$. We present an unsupervised pipeline that uses an iterative elimination approach, removing questions that are certainly non-duplicates and retaining exact or near-duplicate questions. The proposed approach is different from existing paraphrase identification or duplicate detection approaches as it covers different aspects of similarity in a single pipeline with no supervised data. The pipeline proposed is shown in Figure 1. The pipeline consists of the following stages :

\begin{enumerate}
    \item Preprocessing and hierarchical learning taxonomy tagging
    \item Jaccard similarity between questions tagged with similar learning taxonomy.
    \item Named Entity Recognition for computing entity differences.
    \item Overlap of key concepts obtained through concept extraction algorithm and negation detection.
\end{enumerate}

\subsection{Preprocessing and Indexing by Hierarchical Learning Taxonomy}

Given a question $q_{new}$ as input, we preprocess the question, such as sentence level tokenization, removing HTML tags, and non-alphanumeric characters, and removing punctuation marks. The database includes questions asked in high school and belongs to various subjects, including chemistry, physics etc.

Therefore we normalize chemical element abbreviations and symbols to their complete form (Cl → chlorine, pi → $\pi$ etc.) using a dictionary $dict_{sym}$ to ensure consistency resulting in $q_{norm}$.
           \[ q_{norm} =  f_{norm}(q_{new}) \]
           \[ S \gets tokenize(q_{new}) \]
            \[f_{norm} = dict_{sym}[s_i]  \ for \ s_i  \ in \ S\]

After preprocessing the input question, we tag the input question to its standardized hierarchical learning taxonomy of form subject - chapter - topic using the TagRec \cite{tagrec} model. The TagRec approach follows a two-tower transformers-based architecture that aligns the vector subspaces of the input question and the hierarchical learning taxonomy using a contrastive learning approach. We use this trained model to tag our database of questions and $q_{new}$ in a zero-shot setting and index the questions according to the tags.

We extract the subject portion of the taxonomy to which the question belongs and query the complete database to return the candidate set $S_{cand}$ = \{ $q_1$ , $q_2$ , . . . . $q_n$ \} of all the questions in the database that belong to the same subject.  Our dataset primarily consists of questions from the subjects: Physics, Chemistry, Social Science, etc.
For example, the question \textit{How many $\pi$ bonds are present in ferrocene?} belongs to the subject Chemistry.

\subsection{Token level comparison}

After getting the set $S_{cand}$ of the questions belonging to the same subject as from the same hierarchy as the input question $q_{new}$, the model iterates over $S_{cand}$ and checks for the $Jaccard Similarity$ measure.

More formally, Let $q_1$ , $q_2$ be two lists of tokens for the input questions, then Jaccard similarity between these two questions can be calculated as  

\begin{equation}
J(q_1, q_2) = \frac { \# ( q_1 \cap q_2 )} { \# ( q_1 \cup q_2 )}
\end{equation}

%  where \# (A) is the cardinality of set A \\

If the Jaccard similarity ($J(q_{new},q_i)$) between the input question and a question from $S_{cand}$ is less than a certain threshold ($J(q_{new},q_i) < 0.4$) we remove that question from our search space $S_{cand}$. The threshold value of 0.4 was chosen after multiple iterations and validation of the results for the dataset that we worked on.
            \[S_{cand} \gets S_{cand} - q_i \ (if \ J(q_{new},q_i) < 0.4) \]
If the Jaccard similarity is 1 we directly add that question to our exact duplicate question set ($qdup_{exact}$). 

\subsection{NER and comparison}

To further partition $S_{cand}$, we remove questions with different named entities than those in $q_{new}$. For extracting the set of named entities ($NE_q$) of $q_{new}$ we use spaCy, an implementation in Python.
    \[NE_q = NER(q_{new})\]

\noindent Examples :  Who is the CEO of Google ?     →     {'Google': ORG }, Who is the CEO of Apple ?     →     { ‘Apple’ : ORG }

The Named Entity Recognition step performs a sequence labeling task where the noun phrases are tagged with 'PERSON', 'ORG', 'LOC', etc as applicable. Once extracted, the set of entities for $q_{new}$ is compared to the set of entities $NE_i$ for question $q_i$, where i = {1... |$S_{cand}$|}. 
All those questions which have a non-empty difference set between $NE_q$ and $NE_i$ are removed from the search space ($S_{cand}$). 

\[S_{cand} \gets S_{cand} - q_i \ if \ NE_q \cap NE_i \neq \emptyset \]

\subsection{Keyphrase extraction and calculating the overlap}

Following the previous stages of the pipeline, the set $S_{cand}$ = \{  $q_1$ ,  $q_2$ , . . . . $q_n$ \} is left of potential candidates for a duplicate question. The next stage of the pipeline (as shown in Figure 1) is to run an unsupervised method to automatically extract concept terms (keyphrases) from the input question $q_{new}$ into a set $KW_i$. We leverage the EmbedRank algorithm \cite{embedrank} for extracting keyphrases. The proposed approach first extracts candidate phrases using POS tags and projects them and the original question $q_{new}$ to a continuous vector space. It then computes the semantic relatedness between the question and the phrase representations and retrieves the top $k$  keyphrases. For all the questions, we pre-compute the keyphrases and index them. We run a comparison to determine the percentage of keyphrases shared between $KW_i$, and the set of keyphrases extracted for each of the questions in set $S_{cand}$. Questions that have keyphrases sharing score of less than 0.7 (chosen after multiple validations) are eliminated from $S_{cand}$. 

    \[KW_i \gets EmbedRank(q_{new})\]
    \[S_{cand} \gets S_{cand} - q_i \ if \ KW_{share} < 0.7 \]

\subsection{Negation detection}

As a result of the previous steps, the set $S_{cand}$ is much smaller in size and has questions very similar to $q_{new}$. However, we observed that multiple questions with similar verbiage exist though they differ by a negation resulting them having different answers. For example:  \textit{What is an example of a metal ?} and  \textit{What is not an example of a metal ?}

Similar cases might still be left in $S_{cand}$, and hence we check for the difference in negation. We compare $q_{new}$ against each question in $S_{cand}$ and eliminate any questions that may be the negation of $q_{new}$ by ensuring that standard negation constructions, if any, are present in both samples being compared. This ensures that questions with high levels of Jaccard similarity and overlapping keyphrases shares but differing by a single negation are not identified as duplicates (false positives). After this stage, we assign the remaining questions to be duplicates.

\subsection{Related Questions}

The above-mentioned pipeline focuses on a higher recall by sacrificing precision since the problem statement focuses on e-learning platforms being able to rid their database of duplicates. An additional property of our tool is that these platforms can test students on their knowledge of the topic by retrieving related questions which center around the same or similar topics.

Such questions are referred to as \textit{related questions} in this paper and are computed by utilizing the architecture of $all-mpnet-base-v2$ sentence transformers model\footnote{\href{https://huggingface.co/sentence-transformers/all-mpnet-base-v2}}. In the approach, for a duplicate question $q_{dup}$ of an input question $q_{new}$, we find the questions that have embeddings closest to $qdup_{exact}$ or $q_{dup}$ (pre-computed in the database), measured by the cosine similarity between the two embedding vectors and return the 3 closest neighbors.

% More formally, cosine similarity between two vectors $h1$ and $h2$ can be defined as:

% \begin{equation}
% cosine(h_1,h_2) = = \frac { h_1 \cdot h_2 } { \parallel h_1 \parallel \cdot \parallel h_2 \parallel }
% \end{equation}
\begin{figure*}
    \includegraphics[width =0.58\linewidth]{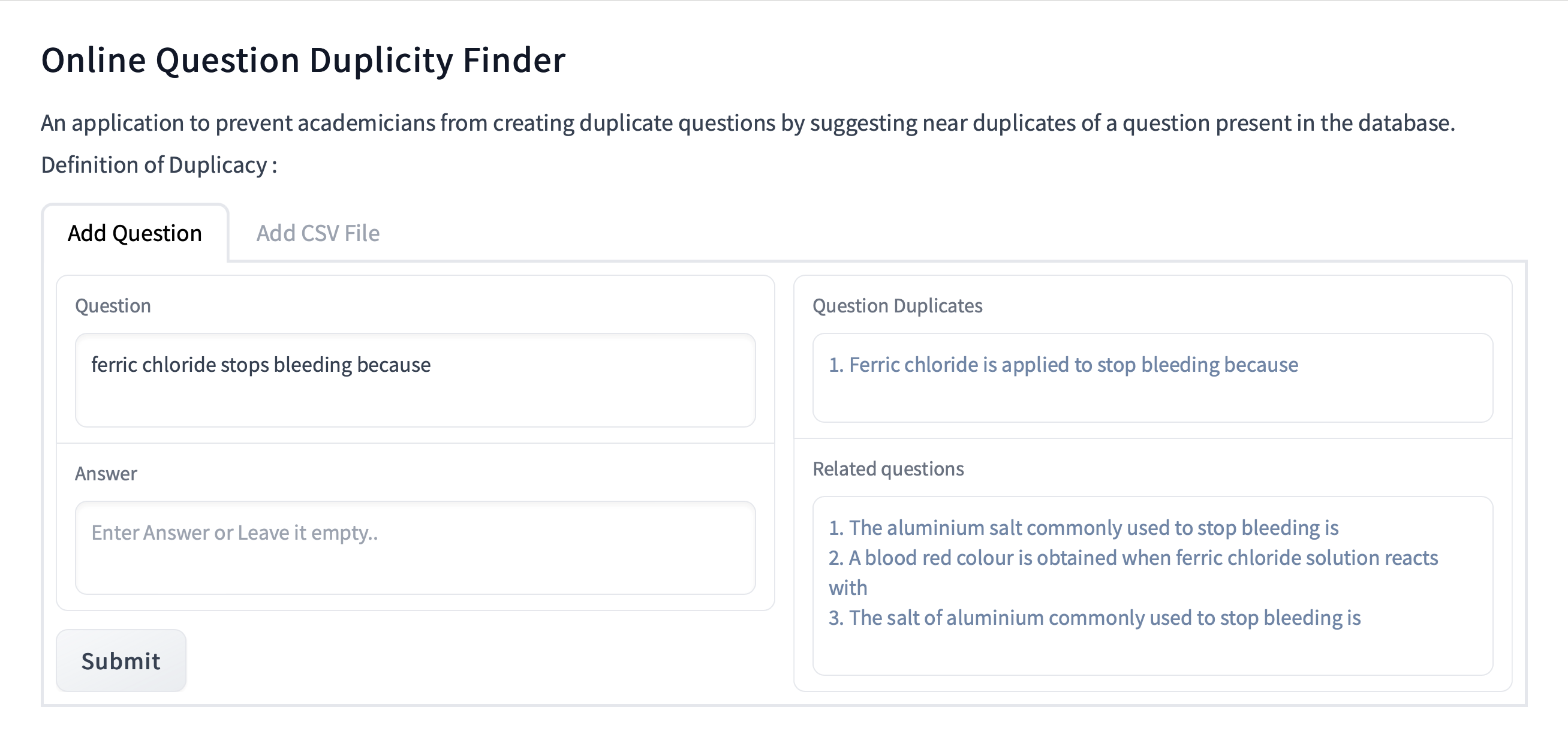}
    \caption{Screenshot of the tool \textit{QDup}}
\end{figure*}

The results demonstrated that nearest neighbors search over a large database gave slow performance, which led us to leverage $ScaNN$ (an efficient \href{https://github.com/google-research/google-research/tree/master/scann}{searching technique} developed by \cite{scann}. The set of embeddings for all the questions in the database is precalculated (using the same $all-mpnet-base-v2$ model) and stored locally for higher efficiency during running.

For every input $q_{new}$, we have $qdup_{exact}$, $qdup$ and related questions $qdup_{rel}$ ($q_{rel}$ is non-empty only if $qdup$ is non-empty).

\section{Demonstration}
We demo our tool from the perspective of it's ability to perform \textbf{near-duplicate detection}, analysis to gauge \textbf{usability} of the tool.
\begin{figure}
    \centering
    \includegraphics[width =0.60\linewidth]{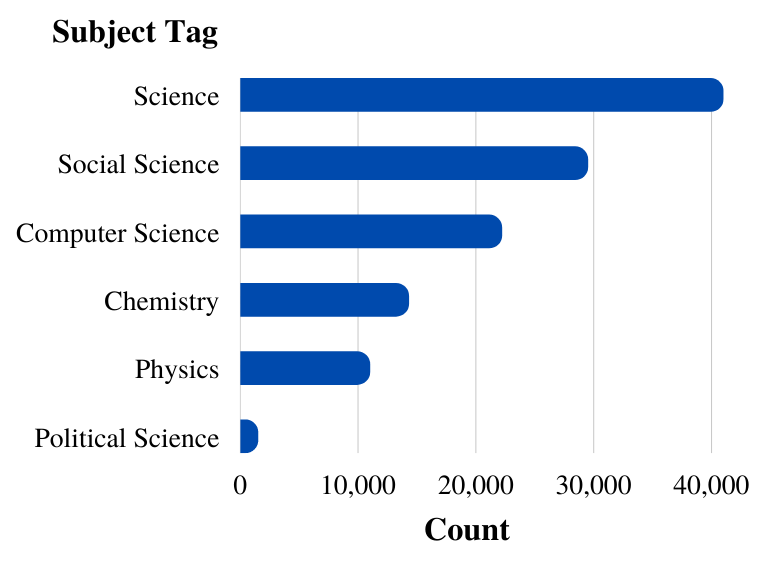}
    \caption{Dataset Statistics}
\end{figure}
% \vspace{-1em}

% \vspace{-1em}
\subsection{Dataset}
The dataset we used consists of 114804 secondary high school questions from the CBSE (India) curated with the help of a leading e-learning platform. The dataset statistics are shown in Figure 3.
\begin{table}[]
    \centering
    \begin{tabular}{c|c}
        \bf Method & \bf Accuracy (\%)  \\ \hline
         \textit{QDup}& \textbf{81.5} \\
         keyphrases based & 76.5 \\
         Closest neighbours & 51.5
    \end{tabular}
    \caption{Performance Evaluation for Duplicate Detection}
    \label{tab:my_label}
\end{table}
\subsection{Evaluation}
The tool was evaluated on a set of 100 input questions by two independent researchers. The tool was provided with 100 random questions across domains and the researchers were requested to label the correct duplicates as 1  or 0 in all other conditions. Similarly, the outputs from other approaches were also provided to the researchers for labeling. These approaches included nearest neighbor search for embeddings extracted with all-mpnet-base-v2 sentence embeddings model and comparison of keyphrases extracted using EmbedRank. We observed a Cohen's kappa of \textbf{0.60}, \textbf{0.72} and \textbf{0.65} in the three scenarios, respectively indicating substantial agreement between annotators. We report the accuracy in Table 1. We observe that the proposed approach \textit{QDup} outperforms classical keyphrases only or vector based nearest neighbor search methods.

\subsection{Tool Ease of Use}
We also conducted a user study with 14 well trained academicians. A screenshot of the tool is shown in Figure 2. We asked the users to rate the tool on a scale of 1-3 (lowest to highest) from aspects of \textit{intuitiveness}, \textit{responsiveness} and \textit{relevance of output}. The \textit{intuitiveness} metric indicates how intuitive and easy to use the interface is without external help. The \textit{responsiveness} measures the response time and \textit{relevance} measures how much the users think the output for the given questions are accurate duplicates. We observed that the average \textit{intuitiveness} score is \textbf{2.46} and average \textit{responsiveness} score is \textbf{2.78}. The average \textit{relevance} score is \textbf{2.68}. We observe that the majority of the users find the tool easy to use.

\section{Conclusion and Future Work}
In this paper, we propose a tool to find duplicates and related questions in a large repository. The proposed approach is resource and time efficient, and the interface is easy to use. In the future, we plan to use the data collected from this tool as weakly supervised data to train a bi-encoder transformer-based model in a contrastive setting to identify duplicate and related questions in one stage. We also plan to explore knowledge distillation and quantization approaches for efficient deployment of the model.

\section{Acknowledgments}
We thank Extramarks and SERB-FICCI for the support.

\bibliographystyle{ACM-Reference-Format}
\bibliography{citations}

\end{document}